\documentclass[final]{cvpr}

\usepackage{booktabs}
\usepackage{times}
\usepackage{epsfig}
\usepackage{graphicx}
\usepackage{amsmath}
\usepackage{amssymb}

\usepackage[pagebackref=true,breaklinks=true,colorlinks,bookmarks=false]{hyperref}

\def\cvprPaperID{4104} 
\def\confYear{CVPR 2021}

\begin{document}

\title{Learning to Aggregate and Personalize 3D Face \\ from In-the-Wild Photo Collection}

\author{Zhenyu Zhang$^{1, 2}$, Yanhao Ge$^{1}$, Renwang Chen$^{1}$, Ying Tai$^{1}$, Yan Yan$^{2}$, Jian Yang$^{2}$, \\
Chengjie Wang$^{1}$, Jilin Li$^{1}$, Feiyue Huang$^{1}$\\
Tencent Youtu Lab, Shanghai, China$^{1}$\\
Nanjing University of Science and Technology, Nanjing, China$^{2}$\\
{\tt\small {zhangjesse@foxmail.com}} \ \  {\tt\small {yanyan, csjyang}@njust.edu.cn}\\
{\tt\small {halege, renwangchen, yingtai, jasoncjwang, jerolinli, garyhuang}@tencent.com}\\
}

\maketitle

\begin{abstract}
   Non-parametric face modeling aims to reconstruct 3D face only from images without shape assumptions. While plausible facial details are predicted, the models tend to over-depend on local color appearance and suffer from ambiguous noise. To address such problem, this paper presents a novel Learning to Aggregate and Personalize (LAP) framework for unsupervised robust 3D face modeling. Instead of using controlled environment, the proposed method implicitly disentangles ID-consistent and scene-specific face from unconstrained photo set. Specifically, to learn ID-consistent face, LAP adaptively aggregates intrinsic face factors of an identity based on a novel curriculum learning approach with relaxed consistency loss. To adapt the face for a personalized scene, we propose a novel attribute-refining network to modify ID-consistent face with target attribute and details. Based on the proposed method, we make unsupervised 3D face modeling benefit from meaningful image facial structure and possibly higher resolutions. Extensive experiments on benchmarks show LAP recovers superior or competitive face shape and texture, compared with state-of-the-art (SOTA) methods with or without prior and supervision $\footnote{Code is available at https://github.com/TencentYoutuResearch/ \\ 3DFaceReconstruction-LAP}$.

\end{abstract}

\section{Introduction}\label{sec::intro}

Monocular 3D reconstruction of human face is a long-standing problem with potential applications including animation, biometrics and human digitalization. It is an essentially ill-posed problem requiring strong assumption, e.g., shape-from-shading approaches \cite{zhang1999shape}. With 3D Morphable Model (3DMM) \cite{blanz1999morphable} proposed, the reconstruction can be achieved through optimization on low-dimensional parameters \cite{romdhani2005estimating, romdhani2003efficient, zhu2014robust}. Recently, deep neural networks are introduced to regress 3DMM parameters from 2D images with supervision~\cite{zhu2016face, richardson20163d, dou2017end, feng2018joint, liu2016joint} or improve 3DMM with non-linearity~\cite{tewari2017mofa, richardson2017learning, tran2018nonlinear, genova2018unsupervised, tewari2018self, zhou2019dense}. Meanwhile, as these single-view guided methods may suffer from 2D ambiguity, other 3DMM-based works are proposed to leverage multi-view consistency~\cite{tuan2017regressing, tewari2019fml, yoon2019self, wu2019mvf, bai2020deep}. While 3DMM provides reliable priors for 3D face modeling, it also brings potential drawbacks: as built from a small amount of subjects (e.g., BFM~\cite{paysan20093d} with 200 subjects) and rigidly controlled conditions, models may be fragile to large variations of identity~\cite{zhu2020beyond}, and have limitations on building teeth, skin details or anatomic grounded muscles~\cite{egger20203d}.

\begin{figure}[t]\centering
\includegraphics[width=0.85\linewidth]{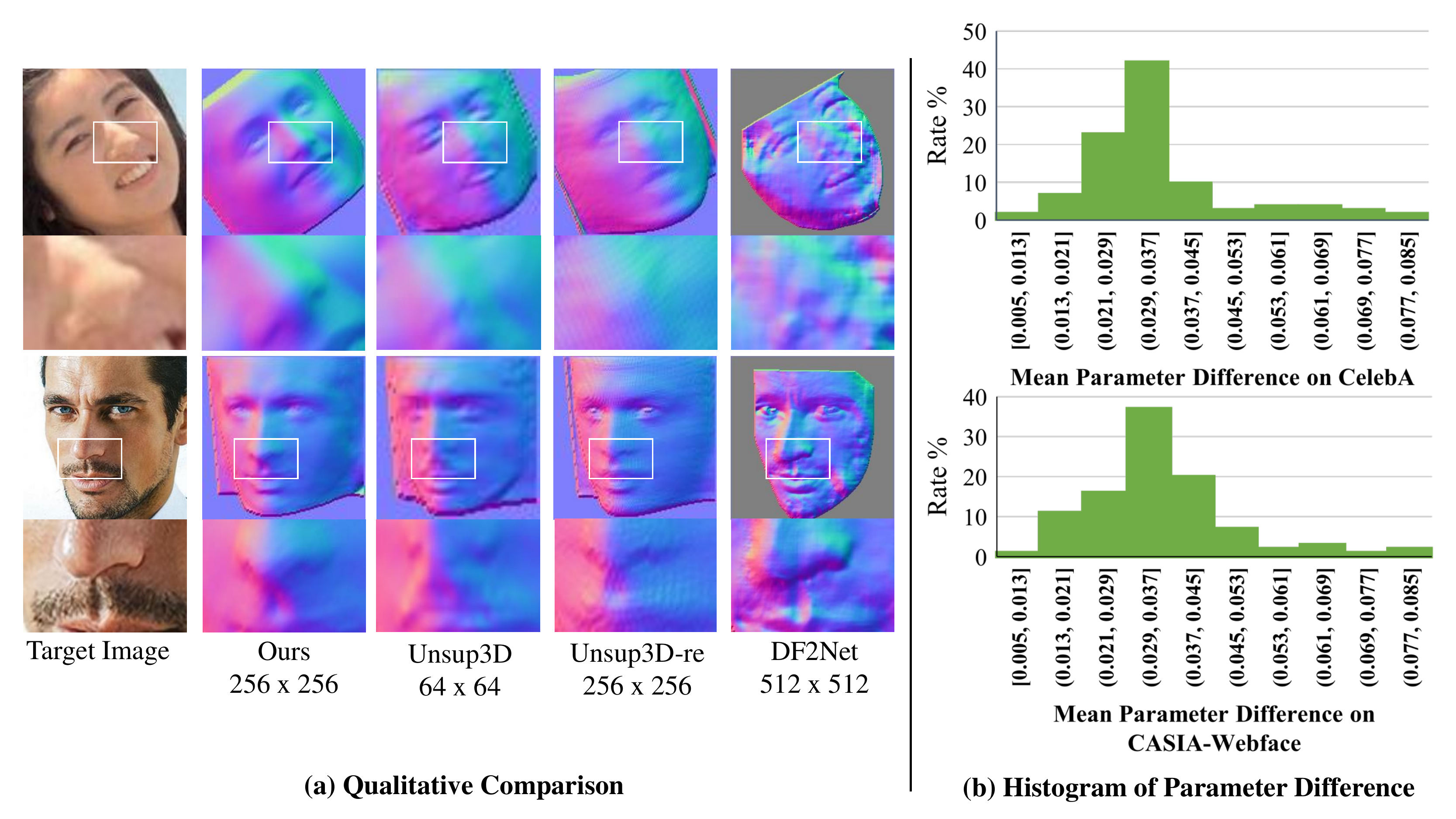}
\caption{(a) Qualitative comparison between our method and Unsup3D~\cite{wu2020unsupervised} and DF2Net~\cite{zeng2019df2net}. Our results show better shape of organs with finer details and less noise.
(b) Distribution of parameter difference in expression basis~\cite{cao2013facewarehouse} between all single-ID image pairs on face dataset~\cite{liu2015deep, yi2014learning}. With the manually computed threshold of 0.04, it reveals that quite amounts (about 70\%) of image pairs have similar expressions and non-rigid difference. Such conclusion inspires us to approximate expressions by mean conditions with reasonable relaxations, and learn a multi-image consistent face without 3DMM prior.
Better showed by zooming in.
}
\vspace{-0.5cm}
\label{fig1}
\end{figure}
Due to the aforementioned limitations, an alternative approach is learning to model 3D face without 3DMM assumption, e.g., regressing face normal or depth directly from an input image~\cite{jackson2017large, trigeorgis2017face, sengupta2018sfsnet, zeng2019df2net, abrevaya2020cross} with ground truth scans and pseudo labels. Despite the efficiency of these approaches, they cannot model facial texture or a canonical view without occlusion.
Recent work Unsup3D \cite{wu2020unsupervised} uses a weakly symmetric constraint to disentangle a face into intrinsic factors and accomplishes the canonical reconstruction in unsupervised manner. In summary, these non-parametric methods predict plausible facial structure via reconstruction or rendering loss~\cite{kato2018neural}. However, without reliable 3DMM prior, they tend to suffer from ambiguity of image appearance. As illustrated in Fig.~\ref{fig1}(a), results of Unsup3D~\cite{wu2020unsupervised} and DF2Net~\cite{zeng2019df2net} have coarse or inconsistent shape of organs. Further, Unsup3D suffers from noise and discontinuity when reproduced for higher-scale reconstruction (Unsup3D-re), which makes the resolution less valuable.
Such phenomenon is due to improper disentangling of albedo, illumination and geometry due to ambiguity of image details and noises as discussed in ~\cite{tran2019towards, lattas2020avatarme, egger20203d}. 
On top of these, we argue that \textit{predicting meaningful and consistent facial structure} is a key point of unsupervised non-parametric 3D face modeling.


To achieve such goal, a better disentangling approach could be: first modeling \textit{basic facial geometry and texture} of an identity, then adding \textit{specific attributes and details} for a target scene.
Actually, basic facial structure is mainly based on bones and anatomic grounded muscles of an identity, which can be enhanced by using 3DMM across unconstrained image set against ambiguous noise~\cite{sanyal2019learning, deng2019accurate, gao2020semi}.
However, multi-image clues are difficult to introduce for non-parametric methods due to 
lack of shape topology. To tackle this problem, we make two assumptions for image collections:
i) Besides the shape, the appearance of an identity due to basic facial structure, like winkles and occlusion of illumination, are similar enough;
ii) Non-rigid shape deformation (mainly about expression) among faces are with limited extent.
The first assumption has been demonstrated by works of~\cite{sanyal2019learning, deng2019accurate, gao2020semi}. For the second one, we compare the expression difference of all image pairs of photo sets in datasets. By using released SOTA 3DMM based model~\cite{deng2019accurate}, we analyse the distribution of mean parameter difference of image pairs on expression basis~\cite{cao2013facewarehouse} in Fig.~\ref{fig1}(b). With the computed similarity threshold 0.04 from manually selected 1k separate image pairs with similar/dissimilar expression, we observe that about 70\% pairs are below the threshold with mild non-rigid difference.
Such conclusion makes it possible to approximate expressions by mean conditions with reasonable relaxations, and learn a multi-image consistent face without 3DMM prior. 
In this paper, we propose a novel Learning to Aggregate and Personalize (LAP) framework for unsupervised non-parametric 3D face modeling. LAP first aggregates consistent face factors of an identity from in-the-wild photo collection, and then personalizes such factors to reconstruct a scene-specific face for a target image of the same ID. Concretely, 
LAP decodes a pair of ID-consistent albedo and depth by adaptively aggregating a global ID code from an image set, and reconstructs a 3D face aligned to each input image based on an estimated specific light and pose. Such aggregation model is optimized by a curriculum learning method with relaxed consistency loss, which helps to overcome large facial variations and lack of pre-defined topology.
Moreover, to personalize a specific face, LAP modifies ID-consistent face through an attribute-refining network for modeling specific attributes and details.
In this way, LAP achieves disentangling of ID-consistent facial structure and scene-specific local details in an unsupervised manner without 3DMM shape assumption. With LAP framework, we manage to model 3D face from arbitrary number of images, or even single image in superior quality and higher resolution than State-of-the-Art (SOTA) methods.

In summary, this paper has contributions in followings:

\textbf{i)} We propose a novel Learning to Aggregate and Personalize (LAP) framework to disentangle ID-consistent and scene-specific 3D face from multi or single image, 
without 3DMM assumptions in fully unsupervised manner.

\textbf{ii)} With a novel relaxed curriculum aggregation method, LAP is able to predict ID-consistent face factors against large facial variations of in-the-wild photo set.

\textbf{iii)} Based on the ID-consistent factors, LAP uses an attribute-refining network to model scene-specific 3D face with less noise and finer details of higher resolutions.


\section{Related Works}
In order to assess our contribution and illustrate contrast between LAP and existing methods, we make a comparison in Table 1. As illustrated, our method faces a more challenging setting, leveraging multi-image consistency from in-the-wild photo set without shape assumption or GT.

\begin{table}[t]
\begin{center}
\footnotesize
\resizebox{.4\textwidth}{!}{
\begin{tabular}{ | l  c  c  c |}
\hline
 & MI-Consistency & Shape Assumption & Supervision \\
 \hline
 \hline
\cite{zhu2016face, richardson20163d, dou2017end, feng2018joint, zhu2020beyond}&
$\times$ & 3DMM & 3DMM paramter \\
\cite{tewari2017mofa, richardson2017learning, genova2018unsupervised, tran2018nonlinear, tran2019towards, tewari2018self, lin2020towards} & $\times$ & 3DMM & I \\
\hline
\hline
\cite{tewari2019fml, wu2019mvf, bai2020deep, shang2020self} &  Constrained & 3DMM & I \\
\cite{sanyal2019learning, deng2019accurate, gao2020semi} & In-the-Wild & 3DMM & I \\
\hline
\hline
\cite{jackson2017large, trigeorgis2017face, abrevaya2020cross, zeng2019df2net} & $\times$ & No & 3DMM parameter, 3D scan, I \\
\cite{sengupta2018sfsnet, sahasrabudhe2019lifting, wu2020unsupervised} & $\times$ & No & I \\
\hline
\hline
Ours & In-the-Wild & No & I \\
\hline

\end{tabular}
}
\end{center}
\caption{Comparison with selected existing method on different settings. Constrained/In-the-wild means the condition of image set, and $\mathbf{I}$ means image. 
}
\vspace{-0.4cm}
\label{table-1}

\end{table}
\textbf{Parametric Method:} With 3DMM~\cite{blanz1999morphable} proposed, 3D face modeling can be formulated in a procedure of parametric optimization~\cite{romdhani2005estimating, romdhani2003efficient, zhu2014robust}. Recently, deep neural networks are introduced to regress 3DMM parameter from input image~\cite{zhu2016face, richardson20163d, dou2017end, feng2018joint, zhu2020beyond} by learning from generated ground truth. With neural rendering approach such as~\cite{kato2018neural}, methods are proposed to leverage image reconstruction loss to train the model in weakly or un-supervised manner~\cite{tewari2017mofa, richardson2017learning, genova2018unsupervised}, or improve 3DMM with more nonlinear feasibility~\cite{tran2018nonlinear, tran2019towards, tewari2018self, lin2020towards, chaudhuri2020personalized}. 
Besides single-view method, multi-view based approaches~\cite{tewari2019fml, yoon2019self, wu2019mvf, bai2020deep, shang2020self} are proposed to model 3D face more robustly. While these methods are based on constrained conditions or video sequence,  they may suffer from limitations for applications. To tackle this problem, methods~\cite{sanyal2019learning, deng2019accurate, gao2020semi} are proposed to use in-the-wild photo collection to improve the robustness of predicted facial shape. Although these approaches have similar motivation to LAP, they are developed based on 3DMM assumption. In contrast, LAP is proposed without such pre-defined topology, thus faces a more challenging problem.

\textbf{Non-Parametric Method:} As an alternative direction, 3D face can also be modeled without 3DMM, e.g., using Shape-from-Shading (SFS) method~\cite{zhang1999shape}. Recently, Sengupta \textit{et al}. propose SFS-Net~\cite{sengupta2018sfsnet} to predict intrinsic factors from input images for modeling 3D faces. With the success of deep neural networks, data-driven methods~\cite{trigeorgis2017face, abrevaya2020cross, jackson2017large, zeng2019df2net} are proposed to directly predict face geometry supervised by real and synthetic ground truth. Despite the efficiency of these approaches, they cannot model 3D geometry of full head or facial textures. A more recent work Unsup3d \cite{wu2020unsupervised} uses weakly symmetric facial constrains to predict light, pose and albedo/depth of canonical view from facial image. Without 3DMM topology, these above non-parametric methods may suffer from ambiguity of image appearance, and predict facial geometry with incorrect details and noise. In contrast, by disentangling a face into ID-consistent and scene-specific factors, LAP models 3D face against such ambiguity and with finer details and structure.

\textbf{Feature Disentangling in Face Reconstruction:} With 3DMM assumptions, methods~\cite{jiang2019disentangled, wu2020leed} can disentangle faces into shapes, expressions and textures. For non-parametric methods, SFS-Net~\cite{sengupta2018sfsnet} and Unsup3d~\cite{wu2020unsupervised} decompose a face into albedo, light, pose and normal. Besides intrinsic decomposition, Deformation AutoEncoder (DAE)~\cite{shu2018deforming, sahasrabudhe2019lifting} disentangles a face into appearance and deformation. Based on such framework, Xing~\textit{et al}.~\cite{xing2019unsupervised} propose a probabilistic method to improve the deformable geometry generation, and Li~\textit{et al}.~\cite{li2019self} leverage videos to urge a better facial action unit. Different from these methods, LAP disentangles a face into global facial structure and scene-specific facial attribute without 3DMM prior from unconstrained photo collection.


\section{Preliminary} \label{sec::pre}

To predict 3D faces without 3DMM assumption, we build our framework based on Unsup3D~\cite{wu2020unsupervised}. Given a face image $\mathbf{I}$, the framework disentangles it into four factors ($d, a, \omega, l$) comprising a canonical depth map $d \in \mathbb{R}_+$, a canonical albedo image $a \in \mathbb{R}^3$, a global light direction $l \in \mathbb{S}^2$ and a viewpoint $\omega \in \mathbb{R}^6$. Each factor is predicted by a separate network which we denote as $\Phi^d, \Phi^a, \Phi^\omega, \Phi^l$.
With these factors, the image $\mathbf{I}$ is reconstructed by lighting $\Lambda$ and reprojection $\Pi$ as follows:
\begin{equation}
\hat{\mathbf{I}} = \Pi(\Lambda(a, d, l), d, \omega).
\label{e1}
\end{equation}
Learning uses a reconstruction loss which encourages $\mathbf{I} \approx \hat{\mathbf{I}}$ with a differentiable renderer~\cite{kato2018neural}. To constrain a canonical view of $d$ and $a$ to represent a full frontal face, the framework uses a weakly symmetric assumption by horizontally flipping:
\begin{equation}
\hat{\mathbf{I}}' = \Pi(\Lambda(a', d', l), d', \omega),
\label{e2}
\end{equation}
where $a'$ and $d'$ are the flipped version of $a, d$, and encourages $\mathbf{I} \approx \hat{\mathbf{I}}'$. To allow probably asymmetric facial region, the framework predicts confidence maps $\sigma, \sigma' \in \mathbb{R}_+$ by $\Phi^\sigma$ and calibrates the loss as follows:
\begin{equation}
\mathcal{L}(\hat{\mathbf{I}},\mathbf{I},\sigma)=-\frac{1}{|\Omega|}\sum{ \ln{\frac{1}{\sqrt{2}\sigma}}\exp{-\frac{\sqrt{2}|\hat{\mathbf{I}}-\mathbf{I}|}{\sigma}}},
\label{e3}
\end{equation}
where $\Omega$ is normalization factor. The flipped version $\mathcal{L}(\hat{\mathbf{I}}',\mathbf{I},\sigma')$ is also calculated. In this way, 3D faces are modeled from images in unsupervised manner without 3DMM assumption. Note that, as Unsup3D extremely depends on single-image appearance, it cannot handle 2D ambiguity such as salient local color difference and noise. In contrast, LAP tackles this problem by further disentangling the face, which will be discussed in the following.

\section{Methodology}

\begin{figure*}[t]\centering
\includegraphics[width=0.7\linewidth]{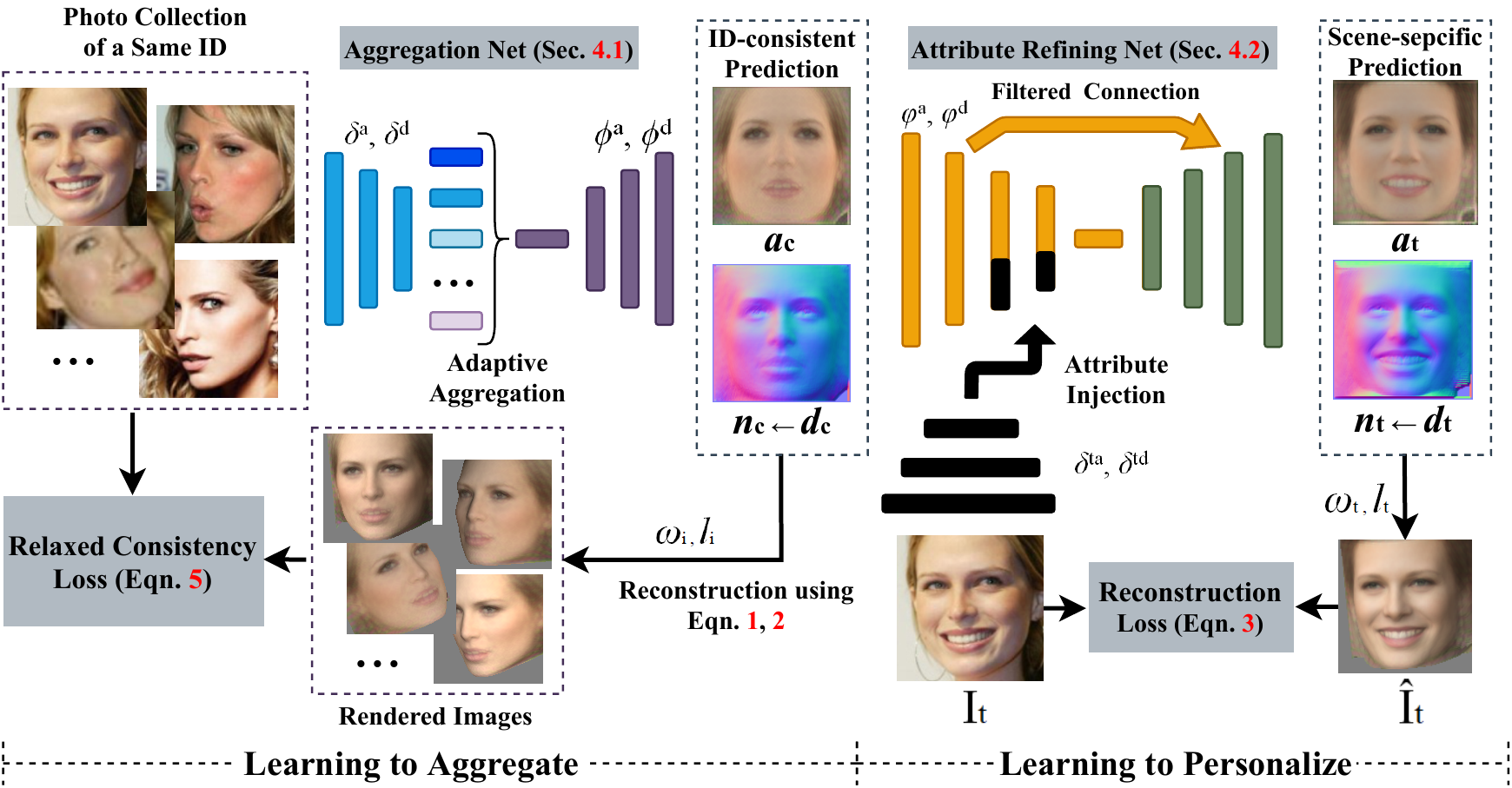}
\caption{Overview of the proposed framework. Training is first conducted on Learning to Aggregate (Sec.~\ref{sec::la}) for modeling ID-consistent face, and then on Learning to Personalize (Sec.~\ref{sec::lp}) for predicting scene-specific face aligned to the target image. $\delta^a, \delta^d$ and $\phi^a, \phi^d$ are the encoder and decoder of albedo/depth aggregation network. $\delta^{ta}, \delta^{td}$ and $\varphi^a, \varphi^d$ are the encoder of attribute injection module and attribute-refining network.
The flipped operation and networks $\Phi^\omega, \Phi^l, \Phi^\sigma$ for predicting pose, lighting and confidence are omitted. 
}
\vspace{-0.4cm}
\label{fig2}
\end{figure*}

In this section, we mainly describe the proposed Learning to Aggregate and Personalize (LAP) 3D face method. With a photo collection of a same identity, our aim is to accomplish a further disentangling: first modeling basic facial geometry/texture based on consistent facial structure,
and then modifying it to personalized attributes and details. As illustrated in Fig.~\ref{fig2}, such disentangling is achieved by two steps: Learning to Aggregate (in Sec.~\ref{sec::la}) and Learning to Personalize (in Sec.~\ref{sec::lp}), without 3DMM priors.

\subsection{Learning to Aggregate} \label{sec::la}

As discussed in Sec.~\ref{sec::intro}, appearance of an identity due to basic facial structure
should be consistent across different images, and image collections contain limited non-rigid variation. Inspired by these facts, we propose depth/albedo aggregation network to adaptively aggregate facial factors from a photo collection and learn ID-consistent geometry/texture, and use such consistent factors to reconstruct each input image. We also propose a curriculum learning approach with relaxed consistency loss to suppress large facial variations for stable learning.

\textbf{Aggregation Network:} As illustrated in Fig.~\ref{fig2}, the aggregation network has a shared encoder $\delta$ across multiple images and a global decoder $\phi$ for predicting consistent face. For modeling albedo and depth, we use two separate aggregation networks denoted as $\Phi^a = (\delta^a, \phi^a)$ and $\Phi^d=(\delta^d, \phi^d)$. Given a photo collection of $N$ images $\{ \mathbf{I}^k_i \}^N_{i=1}$ where $k$ is the index of identity (omitted in the following for simplification), we feed each $\mathbf{I}_i$ into $\delta^a, \delta^d$ to get texture and geometry latent code $\mathbf{x}_i^a, \mathbf{x}_i^d$. Different from the encoder in~\cite{wu2020unsupervised}, to get multi-level information, we downsample the feature of each scale through a convolutional layer with average pooling to a vector, and fuse the vectors through concatenation and a convolutional layer to get $\mathbf{x}_i^a, \mathbf{x}_i^d \in [1,1,c]$.
To learn a global representation of the identity based on $\{\mathbf{x}_i^a, \mathbf{x}_i^d\}^N_{i=1}$, inspired by \cite{liu2017quality, shi2019probabilistic}, we propose an adaptive aggregation method. Due to the quality of $\{ \mathbf{I}_i \}^N_{i=1}$, the importance of each dimension in $\mathbf{x}_i^a, \mathbf{x}_i^d$ which reveals how correlated it is to the identity, is supposed to be different. Hence, we first learn channel-wise weights $\mathbf{w}_i^a, \mathbf{w}_i^d \in [1,1,c]$ to represent the importance, and use softmax function to normalize them
into $\{\bar{\mathbf{w}}_i^a\}^N_{i=1}, \{\bar{\mathbf{w}}_i^d\}^N_{i=1}$. Then the aggregation can be formulated as:
\begin{equation}
\mathbf{x}_c^a = \sum^N_{i=1}{\bar{\mathbf{w}}_i^a \cdot \mathbf{x}_i^a},\ \  \mathbf{x}_c^d = \sum^N_{i=1}{\bar{\mathbf{w}}_i^d \cdot \mathbf{x}_i^d},
\label{e4}
\end{equation}
where $\mathbf{x}_c^a, \mathbf{x}_c^d$ are the combined global ID-code for texture and depth. Compared with naive average fusion, such adaptive aggregation method encourages the ID-correlated features to get larger weights 
in $\bar{\mathbf{w}}_i^a, \bar{\mathbf{w}}_i^d$, thus the fused code $\mathbf{x}_c^a, \mathbf{x}_c^d$ can better represent the consistent features of the identity (see Fig.~\ref{fig5}). Next, we feed $\mathbf{x}_c^a, \mathbf{x}_c^d$ to the decoder $\phi^a, \phi^d$ to get ID-consistent albedo $a_c$ and depth $d_c$. With $\omega_i, l_i$ predicted by $\Phi^\omega(\mathbf{I}_i), \Phi^l(\mathbf{I}_i)$ from each input image, we can reconstruct rendered image $\hat{\mathbf{I}}_i, \hat{\mathbf{I}}_i'$ using Eqns.~\ref{e1},~\ref{e2}. Then multi-image consistency is achieved by calculating Eqn.~\ref{e3} between rendered and original images,
which enhances ID-consistent facial structure in $a_c, d_c$ and suppresses possibly ambiguous noise in each input image.

\begin{figure}[t]\centering
\includegraphics[width=0.7\linewidth]{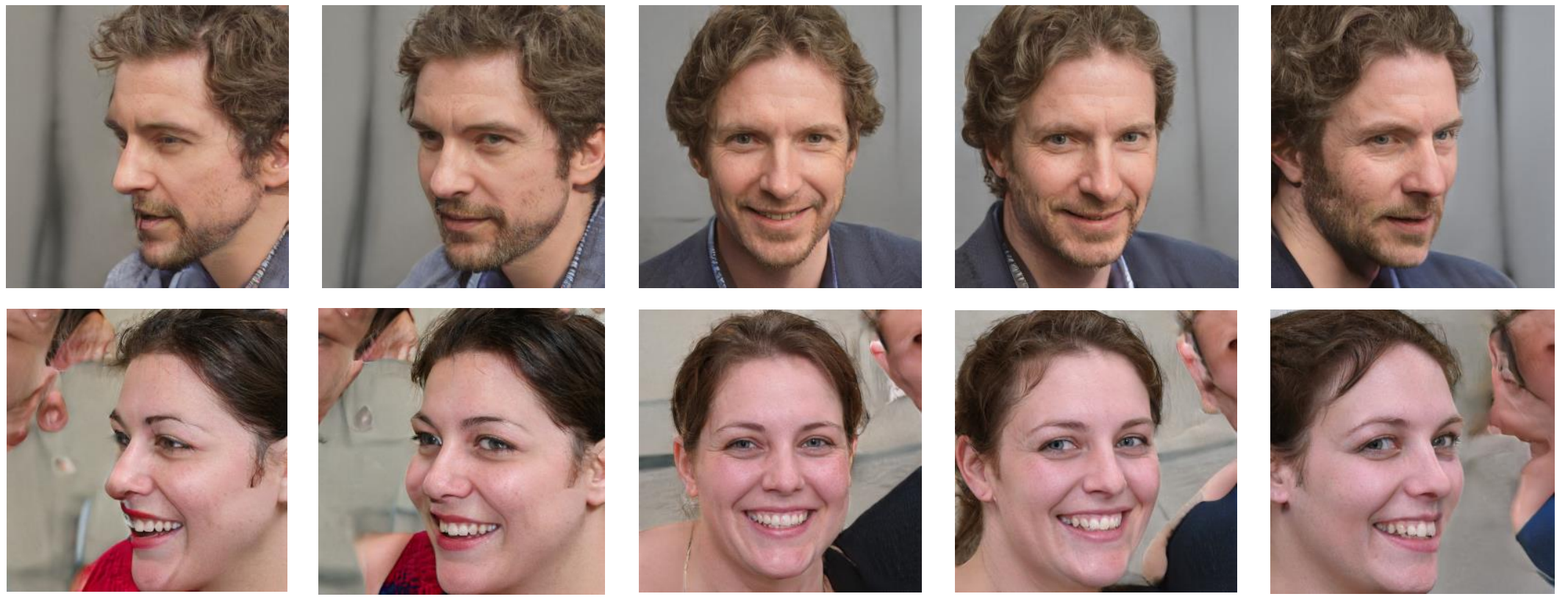}
\caption{Easier samples generated by Interface-GAN~\cite{shen2020interpreting} pretrained on FFHQ dataset~\cite{karras2019style}.
}
\vspace{-0.5cm}
\label{fig3}
\end{figure}

\textbf{Curriculum Learning:}
As illustrated in Fig.~\ref{fig2}, in-the-wild photo collection has different conditions on expression, make-ups, skins and noise,  thus directly using such photo collection for training urges corrupt $a_c, d_c$ or even totally fails to find correspondence (see Fig.~\ref{fig5}) without 3DMM pre-defined topology.
This motivates us to perform a curriculum learning procedure~\cite{bengio2009curriculum, huang2020curricularface}, i.e., training from easier samples to in-the-wild collections. A simple way is to use facial videos such as voxceleb~\cite{nagrani2017voxceleb} in constrained condition, but it may also have drawbacks: the quality of videos and various pose variations cannot be guaranteed. In contrast, we use a controllable GAN~\cite{shen2020interpreting} to generate photo collections. As illustrated in Fig.~\ref{fig3}, the generated samples have different poses and consistent facial structure with high quality and mild variation. 
To keep the ID consistency of generated images, we use Arcface~\cite{deng2019arcface} to filter out samples with cosine-similarity lower than 0.6 compared with the frontal face image. Note that, we do not expect the samples to have exactly same identity, but only similar facial structure which is sufficient enough for model to learn 3D correspondence.
We generate images of 15 different poses from 30k different identities, with resolution of $1024\times1024$ to benefit a better learning.

\textbf{Relaxed Consistency Loss:} To further urge a stable training procedure, we propose a Relaxed Consistency Loss (RCL) to relax the most uncertain facial region in different conditions. We first train a BiSeNet~\cite{yu2018bisenet} on CelebAMask-HQ dataset~\cite{CelebAMask-HQ} for face parsing, and then define the visible facial region and background with constant 1 and 0 to get an attention mask $\mathbf{M}$. As mouth, eyes and brow are more probably inconsistent compared to other regions across different conditions, we set these parts in $\mathbf{M}$ with lower value (e.g., 0.3) to get a new attention mask $\mathbf{M}_{re}$. Then the RCL can be formulated as:
\begin{equation}
\resizebox{.45\textwidth}{!}{
$\mathcal{L}_{RCL}(\hat{\mathbf{I}}_i,\mathbf{I}_i,\sigma_i)=-\frac{1}{|\mathbf{M}_{re}|}\sum{ \ln{\frac{1}{\sqrt{2}\sigma}}\exp{-\frac{\sqrt{2}|\mathbf{M}_{re} \cdot (\hat{\mathbf{I}}_i-\mathbf{I}_i)|}{\sigma_i}}}.$
}
\label{e5}
\end{equation}
Note that, although the confidence $\sigma, \sigma'$ model the uncertainty to some extent, our RCL provides a certain and stronger constraint. Furthermore, for the extremely hard sample such as image with low quality, large occlusion and extreme lighting, the parsing model tends to predict a corrupt facial region which is much smaller than background, which helps us to naturally filter out such samples for stable learning. In this way, the aggregation network can learn valuable consistent feature against inconsistency.

\subsection{Learning to Personalize} \label{sec::lp}

As illustrated in Fig.~\ref{fig2}, while the learned ID-consistent albedo and depth have basic facial structure, they lack details and attribute (e.g., teeth and expression) aligned to a target image $\mathbf{I}_t \in \{ \mathbf{I}_i \}^N_{i=1}$. 
Hence, we propose attribute-refining network to modify $(a_c, d_c)$ to scene-specific $(a_t, d_t)$, which is achieved via attribute injection and filtered connection.

\textbf{Attribute Injection:} 
Attribute injection module has encoder $\delta^{ta}, \delta^{td}$ to extract albedo/depth attribute information from target image $\mathbf{I}_t$, and uses such information to guide the modifying. 
Denote the encoder of attribute refining network as $\varphi^a, \varphi^d$ for encoding $a_c, d_c$ respectively, a direct embedding approach is to fuse features of $\delta^{ta}, \varphi^a$ and $\delta^{td}, \varphi^d$. However, such method brings two problems: Embedding too many low-level features of $\delta^{ta}, \delta^{td}$ which are spatially aligned to $\mathbf{I}_t$ and rendered output $\hat{\mathbf{I}}_t$, urges $\varphi^a, \varphi^d$ to ignore meaningful canonical texture/geometry information and degrades to trivial pure texture auto-encoder;
Embedding too less feature (e.g., only the highest-level feature vector) loses necessary details of predictions. On top of these, we propose a balancing approach with selecting mechanism. Firstly, we only inject features of last three levels (i.e., features with height/width of $8\times8, 4\times4$ and $1\times1$) of $\delta^{ta}, \delta^{td}$; then, we learn a channel-wise weight for each level of feature to select valuable information, and fuse the re-weighted feature with corresponding one in $\varphi^a, \varphi^d$.
In this way, we inject moderate attribute information from $\mathbf{I}_t$ to guide a finer prediction (see Fig.~\ref{fig6}). 

\begin{figure}[t]\centering
\includegraphics[width=0.7\linewidth]{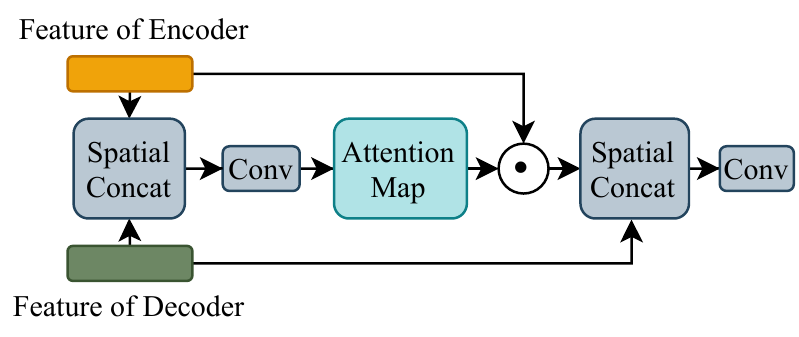}
\caption{Illustration of the proposed Filtered Connection in attribute refining network.
}
\vspace{-0.5cm}
\label{fig4}
\end{figure}
\textbf{Filtered Connection:} To modify $a_c, d_c$ to $a_t, d_t$, the network needs to manipulate facial regions (e.g., mouth, eyes and cheek) which are different to target image, meanwhile suitably keep the structure of unchanged parts. Inspired by works of face editing~\cite{pumarola2018ganimation, zhang2018generative}, we propose a Filtered Connection module to achieve such goal.
As illustrated in Fig.~\ref{fig4}, we firstly combine the features of encoder (yellow block) and decoder (green block) to learn a spatial attention mask $\mathbf{A} \in [h,w,1]$ through convolution and sigmoid function, and then filter the feature by multiplying $\mathbf{A}$ and concatenate it to decoder. In this way, the changed regions of $a_c, d_c$ are suppressed by lower weight in $\mathbf{A}$, while information of unchanged parts is propagated to the decoder.
After getting $a_t, d_t$, we use $\omega_t, l_t$ predicted from $I_t$ to reconstruct the output $\hat{\mathbf{I}}_t$. The learning to personalize framework is then optimized by using Eqn.~\ref{e3} and its flipped version. 

\subsection{Training and Inference} \label{sec::ti}
Training mainly contains three steps: firstly train $\Phi^\omega, \Phi^l, \Phi^\sigma$ and the aggregation network $(\delta^{a}, \phi^a), (\delta^{d}, \phi^d)$ using Eqn.~\ref{e5}; then freeze them and train attribute-refining network using Eqn.~\ref{e3}; finally jointly fine-tune all the networks for finer predictions. During each training stage, we select image set of a same identity with random size (from 1 to 6) as the input of aggregation network. Besides loss functions in Eqns.~\ref{e3},~\ref{e5}, we also use the same perceptual loss as \cite{wu2020unsupervised}. For back-propagation, we use the differentiable renderer~\cite{kato2018neural}.
For inference, using random size of image set or single image (i.e., only the target image) as input, the framework models 3D face for a target image. 

\section{Experiment}
\subsection{Setup}
\textbf{Dataset:} We train our method mainly on the generated synthetic dataset (in Sec.~\ref{sec::la}), CelebA~\cite{liu2015deep} and CASIA-WebFace~\cite{yi2014learning}. To get photo collection of a same identity, we organize CelebA and CASIA-WebFace using ID-labels and keep identities with at least 6 photos. In this way, we get 16k different identities with 600k real face images, and select images of 12k/2k/2k identities as train/val/test set. For evaluation on facial geometry, following \cite{wu2020unsupervised, abrevaya2020cross}, we perform testing on 3DFAW~\cite{gross2010multi, jeni2015dense, zhang2013high, zhang2014bp4d}, BFM~\cite{paysan20093d} and Photoface~\cite{zafeiriou2011photoface} dataset. 3DFAW contains 23k images with 66 3D keypoint annotations, and we use the same protocol as \cite{wu2020unsupervised} to perform testing. For BFM dataset, we use the same generated data released by~\cite{wu2020unsupervised} to evaluate predicted depth maps. Photoface dataset contains 12k images of 453 people with face/normal image pairs, and we follow the protocol of \cite{sengupta2018sfsnet, abrevaya2020cross} for testing. For evaluation on modeled texture, we perform fine-tuning and testing on CelebAMask-HQ~\cite{CelebAMask-HQ} dataset which contains 30k real human facial images with high resolution (1024$\times$1024). We organize it into 24k different identities using groundtruth ID-labels, and randomly select 20k/1k/3k identities as train/val/test set.

\textbf{Implementation Details:} We use the same architecture of $\Phi^\omega, \Phi^l, \Phi^\sigma$ as \cite{wu2020unsupervised} to predict pose, light and confidence. Aggregation and attribute-refining network has the same encoder-decoder backbone as~\cite{wu2020unsupervised} to predict albedo and depth, respectively. As described in Sec.~\ref{sec::ti}, we first train aggregation network and $\Phi^\omega, \Phi^l, \Phi^\sigma$ for 50 epochs on the proposed synthetic dataset, and then continue training on real photo sets for 50 epochs. Then we freeze them and train attribute-refining network for 100 epochs. Finally we fine-tune all the networks for 50 epochs. Training procedure has a batch size of 64 identities and learning rate of $1e-4$ with Adam solver~\cite{kingma2014adam} on one NVIDIA Tesla V100 GPU. We use image set of $128\times128$ as input of aggregation network and get the prediction $a_c, d_c$ of the same size. For scene-specific face, we train different attribute-refining networks according to the size of target image ($64\times64, 128\times128, 256\times256$) and get the prediction $a_t, d_t$ of that size. More details can be found in supp-material.

\textbf{Evaluation Protocol:} As our method can model 3D face for target image using photo set or single image, without special statement, we use single-image results to fairly compare with other methods. For predicted facial geometry, following \cite{abrevaya2020cross, wu2020unsupervised}, we use Scale-Invariant Depth Error (SIDE)~\cite{eigen2014depth} and Mean Angle Deviation (MAD) for evaluating depth and normal. For evaluation on modeled texture, we calculate Structural Similarity Index (SSIM)~\cite{wang2004image} and cosine-similarity of encoded representation of Arcface~\cite{deng2019arcface} between original high-quality images and rendered ones.

\subsection{Ablation Study}
In this section, we perform experiments to analyse the effect of our method. To fairly compare with~\cite{wu2020unsupervised}, we use the exact same network architecture as our method to build the model and train it using the same dataset but without multi-image consistency. We denote such reproduced model as \textbf{Unsup3D-re}. As original model of \cite{wu2020unsupervised} has outputs of low resolution (64$\times$64), Unsup3D-re can make valuable comparison on modeling ability of higher resolution.

\begin{table}[t]
\begin{center}
\footnotesize
\resizebox{.4\textwidth}{!}{
\begin{tabular}{ | c  l  c  c  |}
\hline
   No.  & method   & SIDE ($\times 10^{-2})$ $\downarrow$ & MAD (deg.) $\downarrow$ \\
 \hline
 \hline
(1) & Ours-full & \textbf{0.721}\tiny{${\pm0.128}$}  & \textbf{15.53}\tiny{${\pm1.42}$}  \\
(2) & Unsup3D~\cite{wu2020unsupervised} &  0.793\tiny{${\pm0.140}$}  &  16.51\tiny{${\pm1.56}$}  \\
(3) & Unsup3D-re &  0.785\tiny{${\pm0.152}$}  &  16.44\tiny{${\pm1.63}$}  \\
\hline
\hline
(4) & Ours no-ft &  \textbf{1.102}\tiny{${\pm0.205}$}  &  \textbf{20.75}\tiny{${\pm2.06}$}  \\
(5) & Unsup3D~\cite{wu2020unsupervised} no-ft&  1.295\tiny{${\pm0.233}$}   &  21.84\tiny{${\pm2.56}$}  \\
(6) & Unsup3D-re no-ft &  1.232\tiny{${\pm0.218}$}  &  21.40\tiny{${\pm2.31}$}  \\
\hline
\hline
(7) & w/o curriculum learning &  2.011\tiny{${\pm0.570}$}  &  23.07\tiny{${\pm2.88}$}  \\
(8) & w/o RCL  & 0.738\tiny{${\pm0.135}$}  &  15.66\tiny{${\pm1.50}$}  \\
(9) & w/o adaptive aggregation &  0.764\tiny{${\pm0.142}$}  &  16.21\tiny{${\pm1.76}$}  \\
(10) & w/o filtered connection &  0.725\tiny{${\pm0.139}$}   &  15.33\tiny{${\pm1.58}$}  \\
(11) & w/o attribute injection &  0.750\tiny{${\pm0.157}$}  &   16.01\tiny{${\pm1.20}$}  \\
\hline
\hline
(12) & Ours-full-128 &  0.708\tiny{${\pm0.121}$}  &   15.42\tiny{${\pm1.38}$}  \\
(13) & Ours-full-256 &  \textbf{0.703}\tiny{${\pm0.137}$}  &  \textbf{15.30}\tiny{${\pm1.26}$}  \\
(14) & Unsup3D-re-128 &  0.828\tiny{${\pm0.166}$}  &  18.37\tiny{${\pm1.82}$}  \\
(15) & Unsup3D-re-256 &  0.930\tiny{${\pm0.182}$}  &  19.79\tiny{${\pm1.95}$}  \\
\hline

\end{tabular}
}
\end{center}
\caption{Comparison with Different Baselines and Settings.
}
\vspace{-0.1cm}
\label{table-2}

\end{table}
\textbf{Comparison with Baselines on Geometry:} In Table \ref{table-2} we make comparisons on different baselines and setups. Each model is trained on CelebA and CASIA-WebFace and then fine-tuned on BFM dataset. `No-ft' means the model without fine-tuning. `W/o adaptive aggregation' means using average fusion to fuse latent codes. `W/o attribute injection' means only using feature vector of the highest level to inject target attribute without selecting. Methods with flag `-128' or `-256' mean the output size is $128\times128$ and $256\times256$, while other methods have an output of $64\times64$. As illustrated, Rows (1)-(6) reveal that our method has better ability on specialization and generalization, and outperforms Unsup3D~\cite{wu2020unsupervised} under the same settings. Rows (7)-(11) reveal the effect of each component of our methods, and demonstrate each of them contributes to the final prediction. Rows (1, 3) and (12)-(15) show the ability on modeling faces of higher resolution, which is more challenging due to more complex information on 3D space. The results demonstrate our method has more robust predictions when the resolution increases, while Unsup3D gets obvious performance decline. The above comparisons well support the effectiveness of our method.

\begin{figure}[t]\centering
\includegraphics[width=0.8\linewidth]{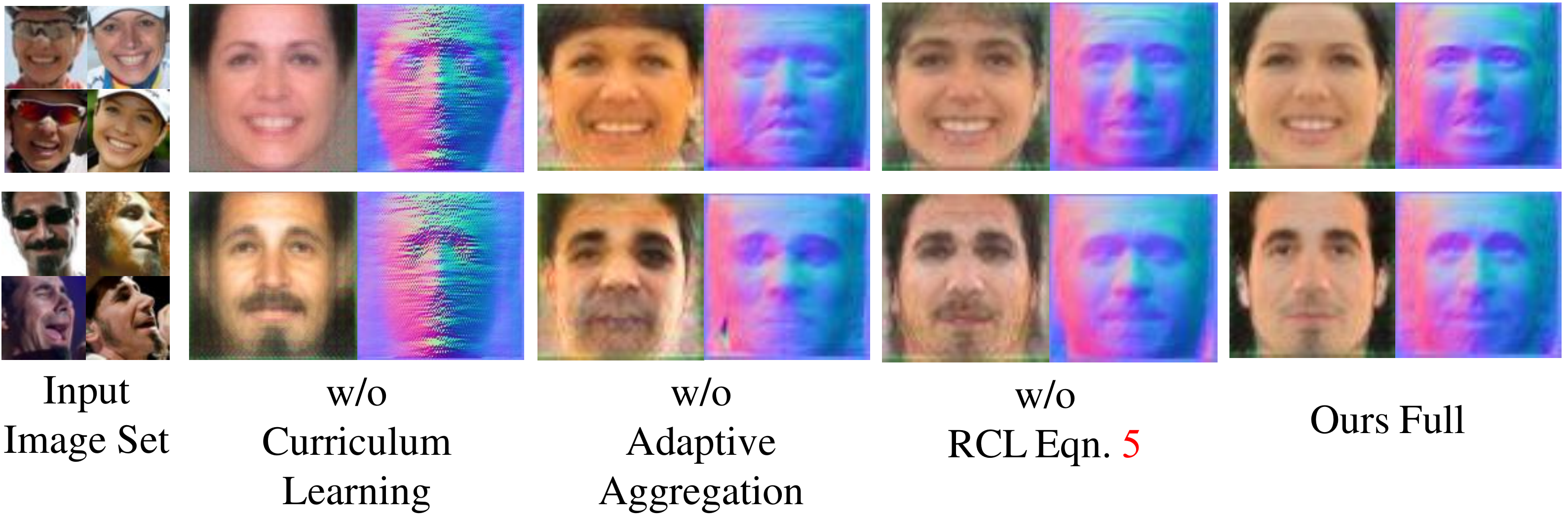}
\caption{Predicted ID-consistent Face under Different Settings. We show canonical albedo and depth to make comparisons.
}
\vspace{-0.1cm}
\label{fig5}
\end{figure}

\begin{figure}[t]\centering
\includegraphics[width=0.8\linewidth]{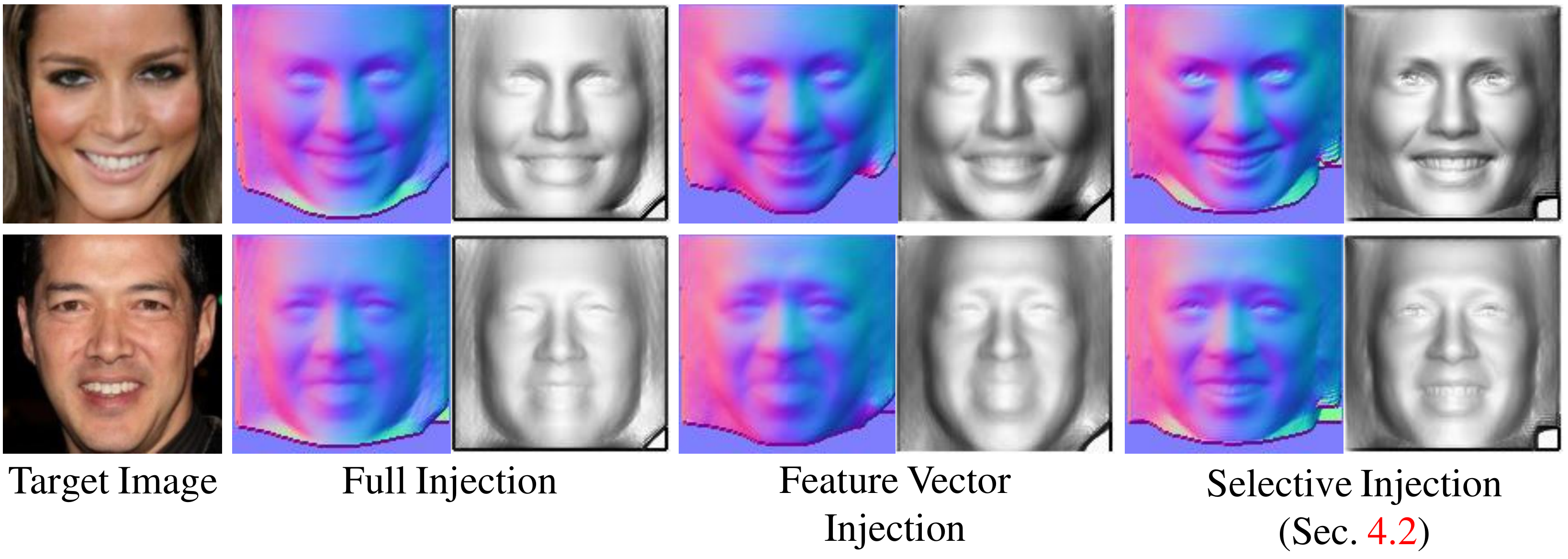}
\caption{Comparison on predicted target facial geometry between baselines and our selective method in attribute injection module. 
}
\vspace{-0.4cm}
\label{fig6}
\end{figure}
\textbf{Qualitative Comparison:} We first compare the predicted ID-consistent face under different settings in Fig.~\ref{fig5}. As illustrated, the aggregation network cannot learn reasonable basic facial geometry without our curriculum learning method. Meanwhile, without adaptive aggregation or RCL, the learned ID-consistent depth and albedo are noisy and suffer from ambiguity. In contrast, our full method can model consistent geometry/texture with feature of the identity from the input image set with obvious higher quality.  Secondly, we compare different injection methods described in Sec.~\ref{sec::lp} in Fig.~\ref{fig6}. As illustrated, injecting full features of target image produces flat and over-smooth geometry, and injecting only the feature vector leads to coarse details. In contrast, our selective injection method models target facial geometry with superior details.

\begin{figure}[t]\centering
\includegraphics[width=0.9\linewidth]{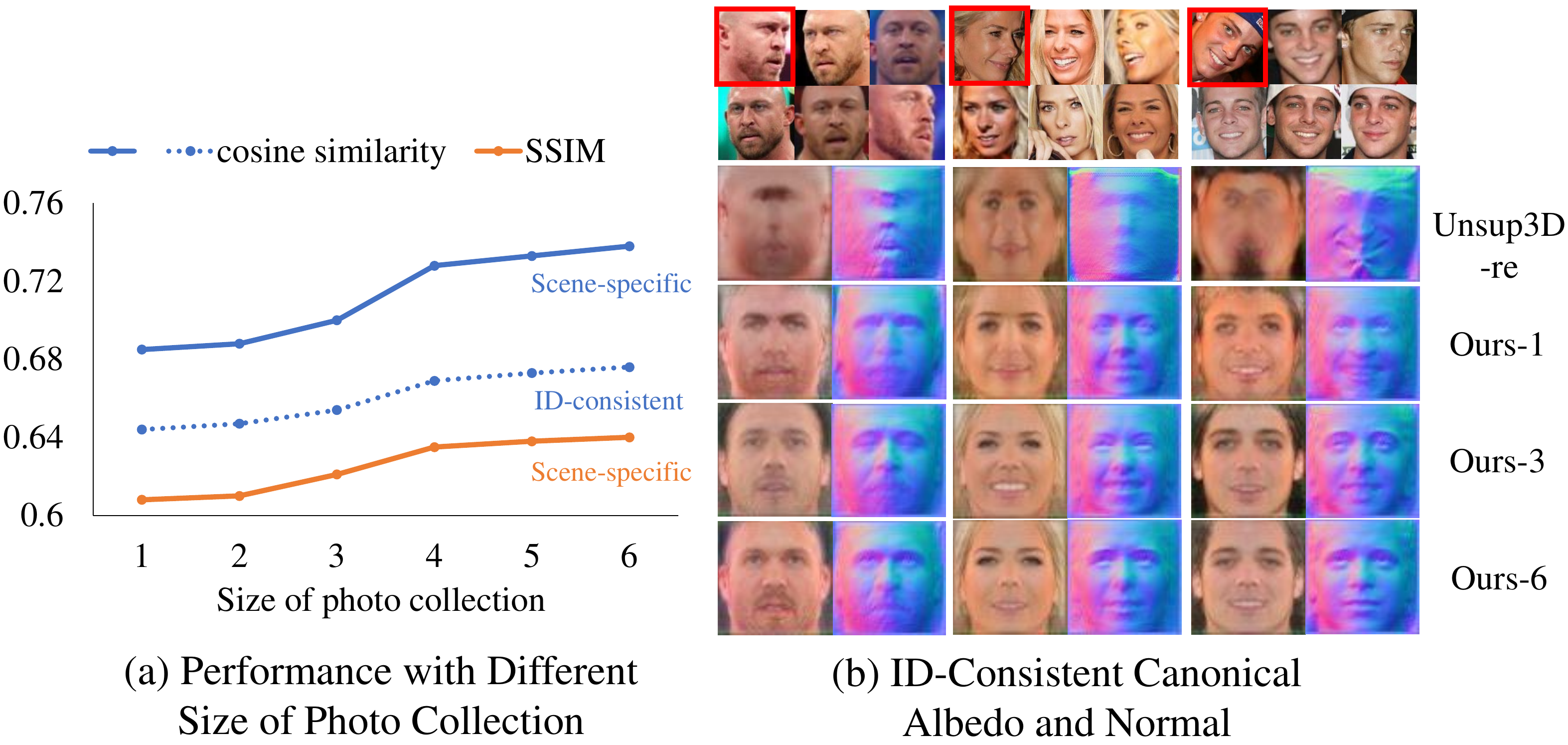}
\caption{ Quality of prediction with different size of photo collection.
(a) The quality of modeled texture on CelebAMask-HQ dataset. (b) Predicted ID-consistent face, the image with red frame is the one used for single-input model (Unsup3D-re and Ours-1).
}
\vspace{-0.2cm}
\label{fig7}
\end{figure}
\textbf{Multi vs. Single:} We analyse the effect of photo collection size, i.e., the $N$ in $\{ \mathbf{I}_i \}^N_{i=1}$. We fine-tune the model on CelebAMask-HQ and evaluate the cosine-similarity and SSIM between $\mathbf{I}_t$ and rendered $\hat{\mathbf{I}}_t$ in Fig.~\ref{fig7}(a). The model predicts target 3D face of $256\times256$, and we calculate SSIM on the same size in facial region. To compute cosine-similarity, we halve the rendered and target image and feed them into the pre-trained Arcface model after alignment for the requirement. As rendered image of ID-consistent face may have different expressions, we only compute cosine-similarity to analyse the predicted ID-feature.
As illustrated, the quality of ID-consistent face increases with the size of image set, and this also contributes to better scene specific prediction. Such phenomenon demonstrates that the aggregated ID-consistent feature is crucial for high-quality modeling. Further,
the quality of modeled texture is similar with 1 or 2 input images, but gets obvious increment with more images. This is due to the sufficiency of complementary information. The increment becomes lighter from 5 to 6, which reveals an potential upper bound. Qualitative results are shown in Fig.~\ref{fig7}(b). Unsup3D-re fails to model reasonable face due to large pose and dis-alignment, while our single-input model (Ours-1) predicts robust results by learning constraint of multi-image consistency. With more input photos, the predictions get better performance.

\begin{table}[t]
\begin{center}
\footnotesize
\resizebox{.35\textwidth}{!}{
\begin{tabular}{ | l  c  c |}
\hline
Method & Depth Corr. $\uparrow$ & Time (ms)\\
\hline
\hline
Ground Truth  &  66  & - \\
AIGN~\cite{tung2017adversarial} (supervised)  &  50.81 & - \\
DepthNetGAN~\cite{moniz2018unsupervised} (supervised)  &  \textbf{58.68} & - \\
\hline
\hline
MOFA~\cite{tewari2017mofa} (3DMM based)  &  15.97 & - \\
DepthNet~\cite{moniz2018unsupervised}  &  35.77 & - \\
Unsup3D~\cite{wu2020unsupervised} (CelebA pre-trained)  &  54.64 & 0.6  \\
Unsup3D-re  &  55.83  & 2.0\\
\hline
\hline

Ours  &  \textbf{57.92} & 2.0 \\
\hline

\end{tabular}
}
\end{center}
\caption{3DFAW keypoint depth evaluation of different methods.
}
\vspace{-0.6cm}
\label{table-3}

\end{table}
\subsection{Comparison with the State-of-the-Art}
\textbf{Analysis on geometry.} We first evaluate the geometry of our predicted 3D face on 3DFAW dataset. To make fair comparison, following \cite{wu2020unsupervised}, we use the 2D keypoint locations to sample our predicted depth and calculate the depth correlation score~\cite{moniz2018unsupervised} on frontal faces. 
As illustrated in Table \ref{table-3}, our method obviously outperforms AIGN, DepthNet and MOFA. For Unsup3D, our method also shows superiority.
Though Unsup3D-re uses our architecture and dataset (CelebA and CASIA-Webface) for training and slightly improves the performance, our method gets further superior result which are closer to fully supervised approach. Inference time of our model is slightly slower than Unsup3D~\cite{wu2020unsupervised}, but our method outperforms Unsup3D-re with the same time burden which demonstrate our implementation is efficient enough.

\begin{table}[t]
\begin{center}
\footnotesize
\resizebox{.4\textwidth}{!}{
\begin{tabular}{ | l  c c c c |}
\hline
 & MAD $\downarrow$ & $<20^{\circ}$ $\uparrow$ & $<25^{\circ}$ $\uparrow$& $<30^{\circ}$ $\uparrow$\\
 \hline
 \hline
Pix2V~\cite{sela2017unrestricted} & 33.9\tiny{${\pm5.6}$} & 24.8\% & 36.1\% & 47.6\% \\
Extreme~\cite{tuan2018extreme} &  27.0\tiny{${\pm6.4}$} & 37.8\% & 51.9\% & 47.6\% \\
FNI~\cite{trigeorgis2017face}  &  26.3\tiny{${\pm10.2}$}  &  4.3\% & 56.1\% & \textbf{89.4}\% \\
3DDFA~\cite{zhu2016face} &  26.0\tiny{$\pm7.2$}  & 40.6\%  &  54.6\%  &  66.4\%  \\
SfSNet~\cite{sengupta2018sfsnet}  &  25.5\tiny{$\pm9.3$}  &  43.6\%  & 57.5\%  &  68.7\%  \\
PRN~\cite{feng2018joint}  &  24.8\tiny{$\pm6.8$}  &  43.1\%  &  62.9\%  &  74.1\%  \\
DF2Net~\cite{zeng2019df2net} (GT) & 24.3\tiny{$\pm5.7$}  &  42.2\%  &  62.7\%  &  74.5\%  \\
D3DFR~\cite{deng2019accurate} & 23.5\tiny{$\pm6.1$}  &  46.1\%  &  61.8\%  &  73.3\%  \\
Cross-Modal~\cite{abrevaya2020cross} (GT)  &  \textbf{22.8}\tiny{$\pm6.5$}  &  \textbf{49.0}\%  &  62.9\%  &  74.1\%  \\
Ours  &  23.0\tiny{$\pm\textbf{5.1}$}  &  48.2\%  &  \textbf{63.1}\%  &  74.9\%  \\
\hline
\hline
SfSNet-ft~\cite{sengupta2018sfsnet}  &  12.8\tiny{$\pm5.4$}  &  83.7\%  &  90.8\%  & 94.5\%  \\
Cross-Modal-ft~\cite{abrevaya2020cross} (GT)  &  \textbf{12.0}\tiny{$\pm5.3$}  &  \textbf{85.2}\%  & 92.0\% & 95.6\% \\
Ours-ft  &  12.3\tiny{$\pm\textbf{4.5}$}  &  84.9\%  &  \textbf{92.4}\%  &  \textbf{96.3}\% \\
\hline

\end{tabular}
}
\end{center}
\vspace{-0.1cm}
\caption{Facial Normal Evaluation on Photoface Dataset.
}
\vspace{-0.4cm}
\label{table-4}

\end{table}
We then evaluate predicted facial geometry on Photoface dataset. Following~\cite{abrevaya2020cross}, we transform our predicted facial depth to normal map and resize it to $256\times256$ in order to compute MAD with ground truth. Results are illustrated in Table~\ref{table-4}, where `-ft' means fine-tuning on Photoface. Our method outperforms most of the approaches and shows good generalization results. For Cross-Modal approach~\cite{abrevaya2020cross}, our method gets competitive results with or without fine-tuning. Note that, the training procedure of~\cite{abrevaya2020cross} utilizes ground truth normal of 3D-scan which is crucial for understanding face geometry, while our model is fully unsupervised thus confronts more challenging conditions. These results demonstrate the effectiveness of our method. 
\begin{figure*}[t]\centering
\includegraphics[width=0.8\linewidth]{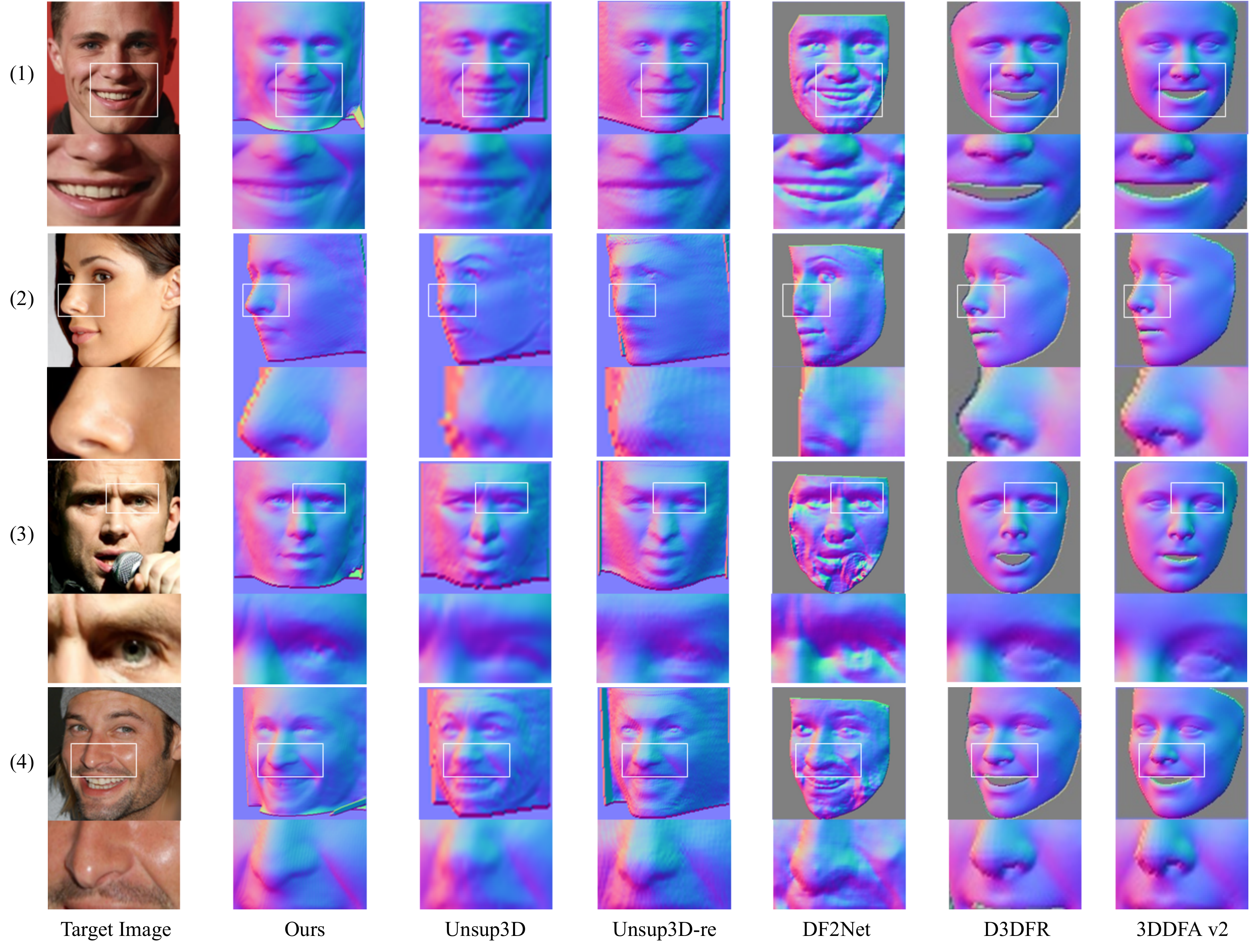}
\caption{Qualitative results on predicted geometry. We compare our method with Unsup3D~\cite{wu2020unsupervised}, DF2Net~\cite{zeng2019df2net}, D3DFR~\cite{deng2019accurate} and 3DDFA v2~\cite{guo2020towards}, and the predictions are reproduced with their released code and pre-trained model. Unsup3D-re means our reproduced model with higher resolution. We use single input for our method to make fair comparison.
}
\vspace{-0.4cm}
\label{fig8}
\end{figure*}

Qualitative results are illustrated in Fig.~\ref{fig8}. Note that Unsup3D has a limited size of output (64$\times$64) and suffers from heavy noise when modeling larger output of 256$\times256$ by our reproduced model (Unsup3D-re), while our method gets obvious superior results on the same resolution.
Compared with non-parametric approaches \cite{zeng2019df2net, wu2020unsupervised}, our method obtains obviously better shape of organs in rows (1), (2) and (3). In rows (2) and (4), our method shows robustness on large pose and artifacts, and suffers from less ambiguity. Compared with 3DMM based methods~\cite{deng2019accurate, guo2020towards}, our results have finer details and recovers better geometric correctness.

\begin{table}[t]
\begin{center}
\footnotesize
\resizebox{.35\textwidth}{!}{
\begin{tabular}{ | l  c  c  |}
\hline
Method & Cosine-similarity $\uparrow$ & SSIM $\uparrow$  \\
\hline
\hline
Unsup3D~\cite{wu2020unsupervised} (64$\times$64) & 0.622 & 0.514  \\
Unsup3D-re (256$\times$256)  &  0.651 & 0.542 \\
D3DFR~\cite{deng2019accurate}  &  0.398  & 0.335  \\
\hline
\hline
Ours ID-consistent (128$\times$128) &  0.643  &  -  \\
Ours (64$\times$64) &  0.695  &  0.618  \\
Ours (128$\times$128) & \textbf{0.697} & 0.620  \\
Ours (256$\times$256) & 0.692 & \textbf{0.623}  \\
\hline

\end{tabular}
}
\end{center}
\caption{Quality of Rendered Image on CelebAMask-HQ.
}
\vspace{-0.4cm}
\label{table-5}

\end{table}
\begin{figure}[t]\centering
\includegraphics[width=0.8\linewidth]{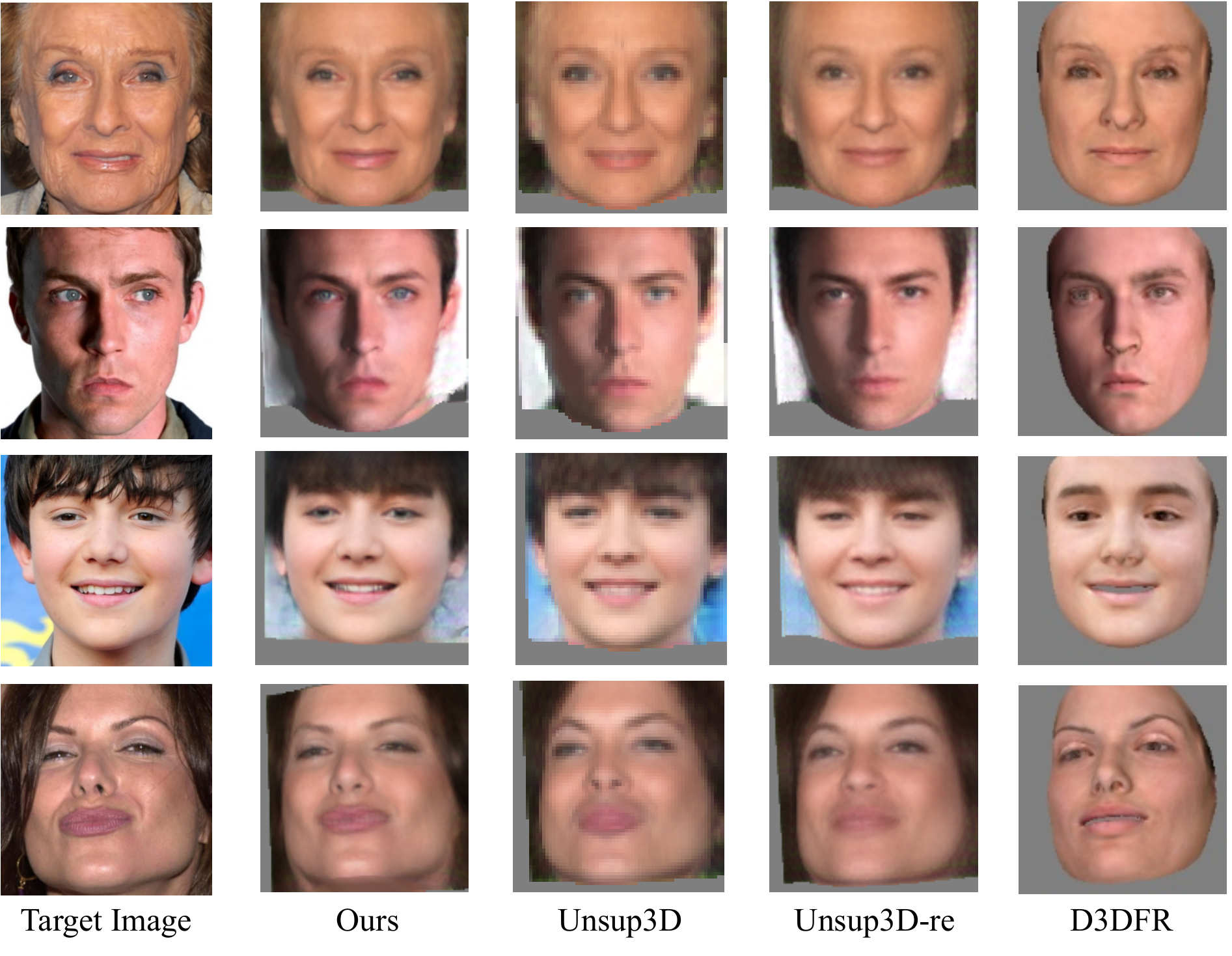}
\caption{Qualitative Comparison on Rendered Image. 
}
\vspace{-0.5cm}
\label{fig9}
\end{figure}
\textbf{Analysis on Texture.} We then analyse our modeled texture on high-quality CelebAMask-HQ dataset. We cover the target image with the modeled texture as the rendered image and use pre-trained Arcface~\cite{deng2019arcface} to compute consine-similarity. SSIM are only computed in facial region. We only compare our single-input results for fairness. As illustrated in Table~\ref{table-5}, the rendered images of our method obtains the best performance. Note that, our ID-consistent predictions also get good consine-similarity, which reveals they aggregate reasonable features of target identities. 
Qualitative results can be viewed in Fig.~\ref{fig9}, and our results have better perceptual quality. 

\section{Conclusion and Future Work}
In this paper we propose a novel Learning to Aggregate and Personalize (LAP) framework for 3D face modeling without 3DMM prior or supervision. Based on statistical conclusion that non-rigid shape deformation is limited in face datasets, LAP adaptively aggregates consistent facial depth and albedo from in-the-wild photo collection, and learns multi-image consistency through a novel curriculum learning method with relaxation. For a face in specific scene, LAP personalizes the consistent face factors by attribute-refining network, improving finer details and attribute. Extensive experiments on benchmarks demonstrate LAP well leverages multi-image consistency and predict superior facial shape and texture. In the future, we may target on modeling 3D face on even higher resolution with reality, and leveraging unconstrained multi-image consistency by explicit algorithm beyond statistical assumption. \footnote{Acknowledgement: We thank the support from members of Tencent Youtu Lab for discussing and improving the ideas.}

{\small
\bibliographystyle{ieee_fullname}
\bibliography{egbib}
}

\end{document}